\documentclass[english]{article}
\usepackage[T1]{fontenc}
\usepackage[utf8]{inputenc}
\usepackage{babel}
\usepackage{verbatim}
\usepackage{float}
\usepackage{amsbsy}
\usepackage{amstext}
\usepackage{graphicx}
\usepackage{esint}
\usepackage[unicode=true,
 bookmarks=false,
 breaklinks=false,pdfborder={0 0 1},backref=section,colorlinks=false]
 {hyperref}

\makeatletter

\providecommand{\tabularnewline}{\\}
\floatstyle{ruled}
\newfloat{algorithm}{tbp}{loa}
\providecommand{\algorithmname}{Algorithm}
\floatname{algorithm}{\protect\algorithmname}

\PassOptionsToPackage{numbers, compress}{natbib}
\usepackage[final]{nips_2017}

\usepackage{url}
\usepackage{booktabs}
\usepackage{amsfonts}
\usepackage{nicefrac}
\usepackage{microtype}
\usepackage{graphics}

\title{Bayesian Conditional Generative Adverserial Networks}

{
\author{
M. Ehsan Abbasnejad\\
 \small{The University of Adelaide} \\
 \texttt{\small ehsan.abbasnejad@adelaide.edu.au}
\And
Qinfeng (Javen) Shi\\
 \small{The University of Adelaide} \\
 \texttt{\small javen.shi@adelaide.edu.au}
\And
 Iman Abbasnejad\\
 \small{Queensland University of Technology \& CMU} \\
\texttt{\small i.abbasnejad@qut.edu.au}
\And 
Anton van den Hengel\\
 \small{The University of Adelaide} \\
\texttt{\small anton.vandenhengel@adelaide.edu.au}, \\
\And
Anthony Dick \\
 \small{The University of Adelaide} \\
 \texttt{\small anthony.dick@adelaide.edu.au}, 
  }
}

\usepackage{babel}
\usepackage{color}
\usepackage{algorithm}
\usepackage{algpseudocode}
\usepackage{subfigure}
\usepackage{wrapfig}

\newcommand{\E}{\mathbb{E}}

\newcommand{\bx}{\mathbf{x}}
\newcommand{\btheta}{{\boldsymbol\theta}}
\newcommand{\bomega}{{\boldsymbol\omega}}

\makeatother

\begin{document}
\maketitle
\begin{abstract}
Traditional GANs use a \emph{deterministic} generator function (typically
a neural network) to transform a random noise input $z$ to a sample
$\bx$ that the discriminator seeks to distinguish. We propose a new
GAN called Bayesian Conditional Generative Adversarial Networks (BC-GANs)
that use a \emph{random} generator function to transform a deterministic
input $y'$ to a sample $\bx$. Our BC-GANs extend traditional GANs
to a Bayesian framework, and naturally handle unsupervised learning,
supervised learning, and semi-supervised learning problems. Experiments
show that the proposed BC-GANs outperforms the state-of-the-arts. 
\end{abstract}

\section{Introduction}

Generative adversarial nets (GANs) \cite{GoodfellowPouget-AbadieMirzaEtAl2014}
are a new class of models developed to tackle unsupervised learning
long standing problem in machine learning. These algorithms work by
training two neural networks \emph{generator} and a \emph{discriminator}–to
play a game in a minimax formulation so that the generator network
learns to generate fake samples to be as ``similar'' as possible
to the real ones. The discriminator on the other hand learns to distinguish
between the real samples and the fake ones. From an information-theoretic
view, discriminator is a measure that learns to evaluate how close
the distribution of the real and fake samples are \cite{ArjovskyChintalaBottou2017,NowozinCsekeTomioka2016}.
Generator network is a deterministic function that transforms an input
noise to samples from the target distribution, e.g. images.

Original GAN algorithm has been extended to conditional models where
in addition to the input noise for the generator, an attribute vector
such as the label is also provided. This helps with generating samples
from a particular class and adding this vector to any layer of the
generator network will effect the performance. In this paper, we propose
to replace the deterministic generator function with a stochastic
one which leads to simpler and more unified model. As shown in Figure
\ref{fig:intro_graph} we omit the need for a random vector in the
input. Furthermore, generator network learns to utilize the uncertainty
in it for generating samples from a particular class that leads to
activation of certain weights for each class.

This representation of uncertainty in the generator (which is easily
extended to the discriminator as well) allows us to introduce \emph{Bayesian
Conditional GAN} (BC-GAN)–a Bayesian framework for learning in conditional
GANs. By integrating out the generator and discriminator functions
we bring about all the benefits of Bayesian methods to GANs: representing
uncertainty in the models and avoiding overfitting. We use dropout,
both Bernoulli and Gaussian, to build our model.

Since training the GANs involve alternating training of the generator
and discriminator network in a saddle-point problem, the optimization
is very unstable and difficult to tune. We believe utilizing Bayesian
methods, where in a Monte Carlo fashion we average over function values
will help with stabilizing the training.

We make the following contributions: 
\begin{itemize}
\item We propose a conditional GAN model that naturally handles supervised
learning, semi-supervised learning and unsupervised learning problems. 
\item Unlike traditional methods using a random noise variable for the generator,
we use a random function that takes deterministic input (see Figure
\ref{fig:intro_graph}). This allows us to utilize the uncertainty
in the model rather than the noise in the input. 
\item We provide a Bayesian framework for learning GANs that capture the
uncertainty in the model and the samples taken from the generator.
Since Bayesian methods integrate out parameters, they are less susceptible
to overfitting and more stable. 
\item We incorporate \emph{maximum mean discrepancy} (MMD) measure to GANs
different from what has been exploited in GANs to further improve
the performance. 
\end{itemize}

\section{Bayesian Conditional GAN}

Let $S=\left\lbrace {(\bx_{1},y_{1}),\ldots,(\bx_{n},y_{n})}\right\rbrace $
and $S'=\left\lbrace {(\bx'_{1},y'_{1}),\ldots,(\bx'_{n},y'_{n'})}\right\rbrace $
be the set of real data and the set of fake data respectively with
$\bx_{i}\in\mathbb{R}^{N\times N}$ and $y_{i}\in\lbrace{1,\ldots,K}\rbrace$.
This may seem to work for supervised learning only, but it actually
works for semi-supervised and unsupervised learning problems for GANs.
In the supervised learning setting, $K$ is the number of classes
for all data. In the semi-supervised setting where we have some unlabelled
data, we can augment the real set $S$ by assigning the unlabelled
data with label $y=K+1$. In the unsupervised learning setting (i.e.
all we have is unlabelled data), the real set is labeled with $y=1$
and fake ones are $y=0$.

\begin{figure}
\centering\subfigure[Original GAN]{\includegraphics[scale=0.3]{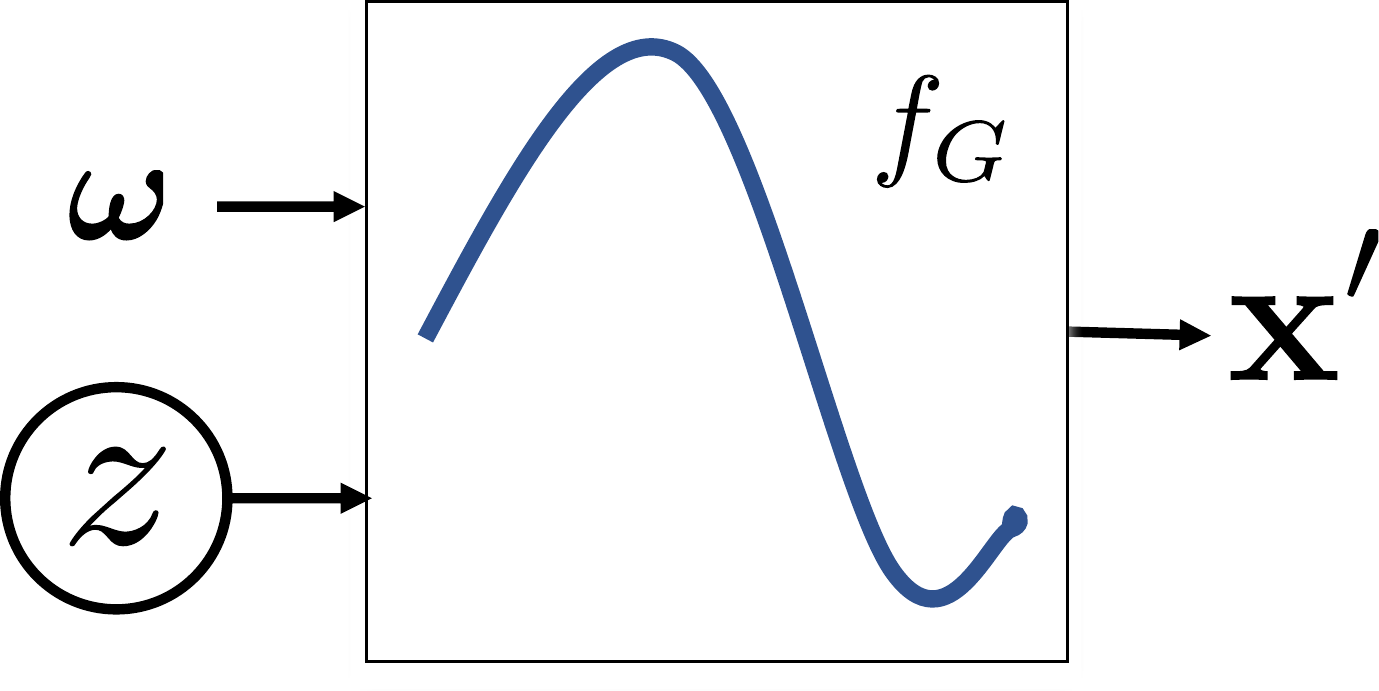}}\qquad\subfigure[BC-GAN]{\includegraphics[scale=0.3]{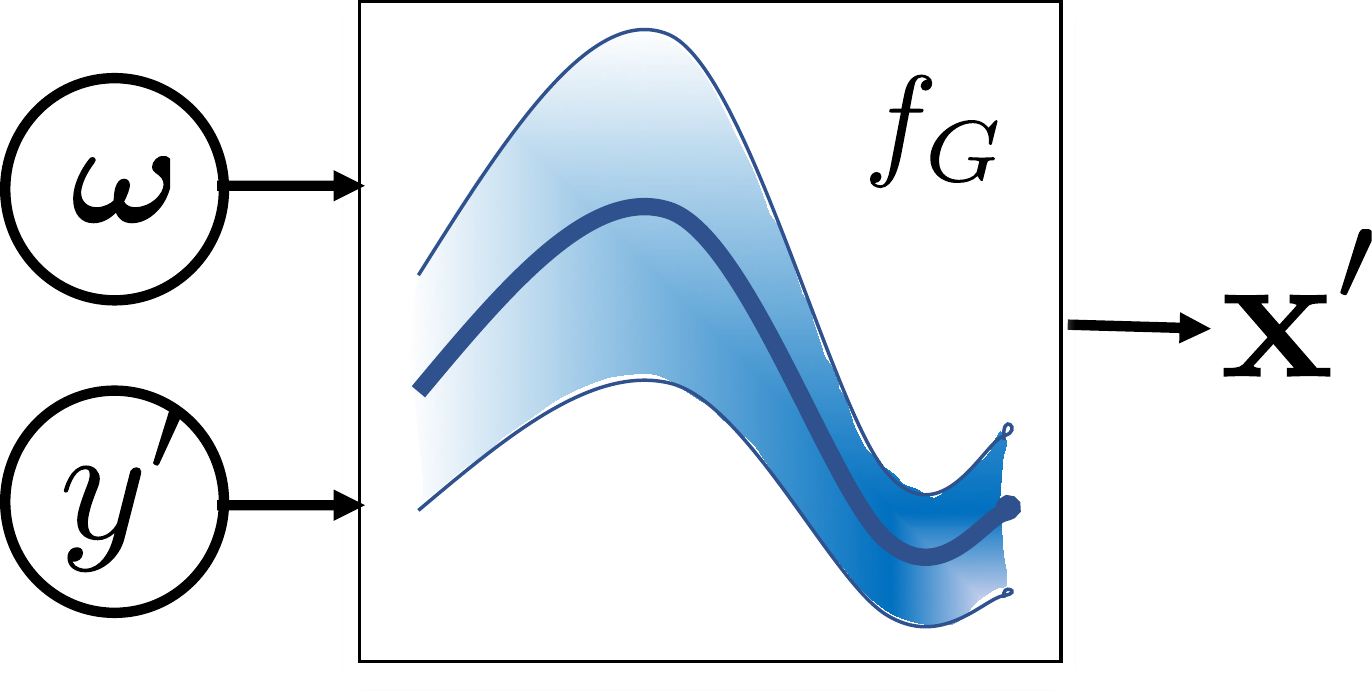}}\label{fig:intro_graph}
\caption{Difference between Original GAN and the Bayesian GAN proposed in this
paper. In our approach, $\bomega$ as the parameter of the generator
in original GAN is a random variable itself. Moreover, $y'\in\mathcal{Y}$
is a deterministic label variable feed into the generator. Each sample
of the data is generated from a sample of the generator function.}
\end{figure}

In many GANs such as Wasserstein GAN \cite{ArjovskyChintalaBottou2017},
the generator is a function that transforms a random noise input to
a sample that the discriminator seeks to distinguish (see Figure \ref{fig:intro_graph}).
In our approach on the other hand, we model the generator as a \emph{random
function} $f_{G}$ that transforms a deterministic input $y'$ to
a sample $\bx$ whose distribution resembles the distribution of real
data (see Figure \ref{fig:intro_graph} Bottom).

We define the distribution of a set of generated samples from the
generator as {\footnotesize{}
\begin{eqnarray*}
p(S'|f_{D}) & \propto & \int p(\bomega)\int p(f_{G}|\bomega)\prod_{i=1}^{n}p(f_{D}(f_{G}(y_{i}'),y_{i}'))df_{G}d\bomega,\\
p(S'|\bomega,f_{D}) & = & \int p(f_{G}|\bomega)\prod_{i=1}^{n}p(f_{D}(f_{G}(y_{i}'),y_{i}'))df_{G}
\end{eqnarray*}
}where $\bomega$ is the parameter/weights of the generator network,
and $p(\bomega)$ is the prior on $\bomega$. $f_{D}$ is the discriminator
function that measures the compatibility of input $\bx$ and output
$y$.


Similarly, we define the distribution of a set of real samples from
the the discriminator as {\footnotesize{}
\begin{eqnarray*}
p(S) & = & \int p(\btheta)\int p(f_{D}|\btheta)\prod_{i=1}^{n}p(f_{D}(\bx,y_{i}))df_{D}d\btheta.\\
p(f_{D}|S) & = & \int p(\btheta)p(f_{D}|\btheta)p(S|f_{D})d\btheta=\int p(\btheta)p(f_{D}|\btheta)\prod_{i=1}^{n}p(f_{D}(\bx),y_{i})d\btheta\\
p(f_{D}|S,\btheta) & \propto & p(f_{D}|\btheta)p(S|f_{D})
\end{eqnarray*}
}where $\btheta$ is the parameter/weights of the discriminator network.
These resemble Gaussian processes (GPs) for classification problems.
In fact, it was shown that using the dropout in a discriminator type
network resembles the posterior estimation in the GPs \cite{GalGhahramani2015}.

The advantage of using the Bayesian approaches in inference of the
parameters is that we include model uncertainty in our approach and
will be better equipped to tackle the convergence problem with GANs.
This is because using weights from the posterior and taking advantage
of the functional distribution, the learner can navigate better in
the complicated parameter space. We observe that this helps with general
GAN's problem of not reaching the saddle point due to the alternation
optimization in both generator and discriminator.

To estimate the expectations and perform inference we turn to commonly
used Monte Carlo methods. In the following we will discuss and experiment
with two of these methods. One is Markov Chain Monte Carlo and the
other is Gradient Langevin dynamics. Due to uncertainty in the model
and the randomness of the generator function, we observed that multiple
rounds of generator update performs better in practice. In other words,
we sample generator more often than updating the discriminator.

With the definitions of the distributions of generator and discriminator,
we now show how to learn them below. 

\begin{figure}
\begin{centering}
\centering\includegraphics[scale=0.3]{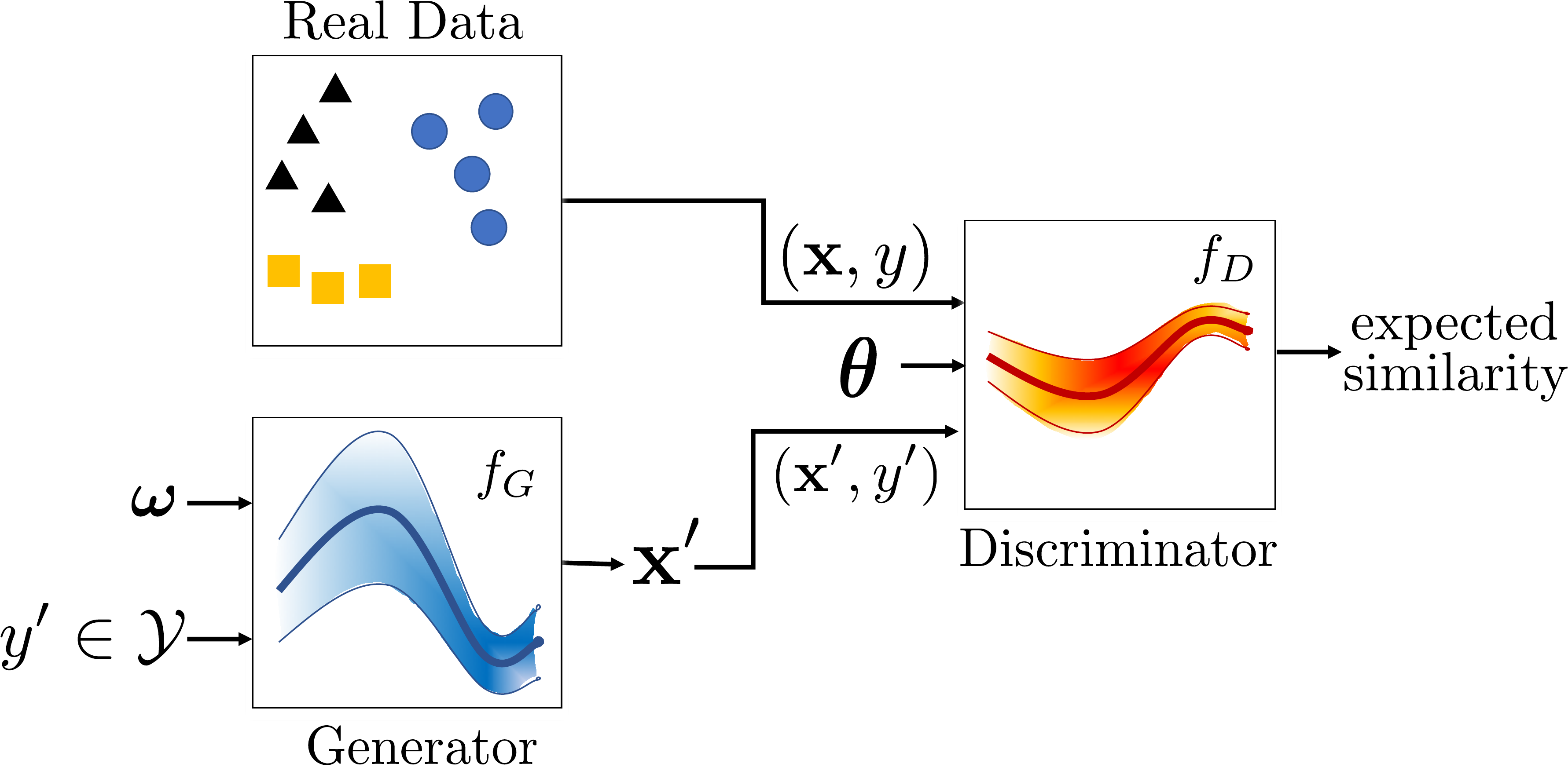} 
\par\end{centering}
\caption{An illustration of the role of the generator and discriminator in
our approach. Rather than }
\label{fig:map} 
\end{figure}

\subsection{MAP Estimate and Sampling}

A simple approach for using the distribution of the transformation
function and the uncertainty of the discriminator is to sample from
the weight distribution and then perform the GAN updates. For this,
we sample the functional values of the generator and discriminator
and minimize our network's loss accordingly. This approach is on par
with performing \emph{Thompson sampling} used in sequential decision
making where an agent picks an action iteratively to minimize an expected
loss within a Bayesian framework. Here, generator and discriminator
play Thompson sampling against each other where at each iteration
based on the current observations (samples from the fake and real
data) the distribution of discriminator (reward function) is updated
. 

{\small{}
\begin{eqnarray}
\min_{\btheta} &  & \E_{f_{D}\sim p(f_{D}|S,\btheta)}\E_{(\bx,y)\sim p_{\text{data}}}[\ell_{D}(f_{D}(\bx),y)]\label{eq:gan_obj}\\
 &  & \qquad\qquad\qquad-\E_{f_{D}\sim p(f_{D}|S,\btheta)}\E_{(\bx',y')\sim p_{\text{fake}}}[\ell_{G}(f_{D}(\bx'),y')]\qquad\text{Discriminator Inference}\nonumber \\
\min_{\bomega} &  & \E_{f_{D}\sim p(f_{D}|S,\btheta)}\Bigg[\E_{(\bx',y')\sim p_{\text{fake}}}[\ell_{G}(f_{D}(\bx'),y')]]\nonumber \\
 &  & \qquad\qquad+\lambda\E_{(x,y)\sim p_{\text{data}}}\E_{(\bx',y')\sim p(S'|\bomega)}[\Delta_{f_{D}}((\bx,y),(\bx',y'))]\Bigg]\qquad\text{Generator Inference}\nonumber 
\end{eqnarray}
}{\small \par}

Here $p_{\text{data}}$ is the true underlying distribution of the
data, and $p_{\text{fake}}$ is the fake distribution of the data
represented by the generator. $\ell_{D}$ is the loss function of
the discriminator network, and $\ell_{G}$ is the loss of the generator
network. $\Delta_{f_{D}}(\bx,y),(\bx',y'))$ describes the discrepancy
of $(\bx,y)$ and $(\bx',y')$. The overall framework of our method
is shown in Figure \ref{fig:map}. 

Since Monte Carlo is an unbiased estimator of the expectations, we
perform MAP on the parameters of the function and then sample the
functions themselves as follows and we call this approach MAP-MC,
{\footnotesize{}
\begin{eqnarray*}
\min_{\btheta} &  & L_{D}(\btheta)\\
 &  & \text{where }L_{D}(\btheta)=\frac{1}{n\times m}\sum_{i}\sum_{j}\ell_{D}(f_{D}^{(i)}(\bx_{j}),y_{j})-\frac{1}{n'\times m'}\sum_{j}\ell_{G}(f_{D}(\bx_{j}'),y_{j}')\\
 &  & \qquad\qquad f_{D}^{(i)}\sim p(f_{D}^{(i)}|S,\btheta)\\
\min_{\bomega} &  & L_{G}(\bomega)\\
 &  & \text{where }L_{G}(\bomega)=\frac{1}{n'\times m'}\sum_{j}\ell_{G}(f_{D}(\bx_{j}'),y_{j}')+\frac{\lambda}{m'}\Delta_{f_{D}^{(i)}}(S,S')\\
 &  & \qquad\qquad f_{D}^{(i)}\sim p(f_{D}^{(i)}|S,\btheta),S'\sim p(S'|f_{G}^{(i)}),f_{G}^{(i)}\sim p(f_{G}^{(i)}|\bomega)
\end{eqnarray*}
}{\footnotesize \par}


\subsection{Full Bayesian using Stochastic Gradient Langevin Dynamics}

Another way to perform inference in our GAN model, is to employ \emph{stochastic
gradient Langevin dynamics}. Inspired by Robbins-Monro algorithms,
this MCMC approach is proposed to perform more efficient inference
in large datasets. In principle, Langevian dynamics takes the updates
of the parameters in the direction of the maximum a posteriori with
injecting noise so that the trajectory covers the full posterior.
Thus, updating the discriminator and generator network by adding noise
to the gradient of the model updates. This is particularly used when
the losses $\ell_{D},\ell_{G}$ give rise to distributions e.g. in
case of softmax loss.

Langevin dynamics allows us to perform the full Bayesian inference
on the parameters with minor modifications to the pervious approach.
To use Langevin dynamics, we update the parameters with added Gaussian
noise to the gradients, i.e. {\footnotesize{}
\begin{eqnarray}
\btheta & = & \btheta-\frac{\eta_{t}}{2}\times\bigg(\sum_{j}\nabla_{\btheta}L_{D}^{(j)}(\btheta)\bigg)+r_{t}\nonumber \\
\bomega & = & \bomega-\frac{\eta_{t}}{2}\times\bigg(\sum_{j}\nabla_{\bomega}L_{G}^{(j)}(\bomega)\bigg)+s_{t}\nonumber \\
 &  & r_{t}\sim\mathcal{N}(0,\eta_{t}),s_{t}\sim\mathcal{N}(0,\eta_{t})\label{eq:langevin_update}
\end{eqnarray}
}This added noise will ensure the parameters are not only traversing
towards the mode of the distributions but also sampling them according
to their density. In practice, to improve convergence of our GAN model,
we use a smaller variance in noise distribution.

\begin{algorithm}
\begin{algorithmic}[1] 
\Require  	
	\Statex {$\eta: \text{learning rate}$} 	
	\Statex {$\lambda: \text{MMD regularizer}$} 	
	\Statex {$\pi_D, \pi_G: \text{dropout probability for discriminator and generator respectively}$}
	\Statex {$\sigma_D, \sigma_G: \text{standard deviation of weight prior for discriminator and generator respectively}$} 
	\Statex \State Initialize $\btheta, \bomega$ randomly 
\While{not converged} 	
	\State{$S=$Sample a batch $\{(\bx_i,y_i)\}_{i=1}^n$ from the real data distribution} 	
	\For {$j=1\to m$} 	
		\State{Sample $\boldsymbol{\alpha}\sim\text{Bernoulli}{(\pi_D)}$} 	
		\State{Sample $\boldsymbol{\beta}\sim\mathcal{N}(0, \sigma_D^2\mathbf{I})$} \Comment{According to dimenstions of $\bomega$} 	
		\State $\tilde{\btheta}=\btheta\odot\boldsymbol\alpha+\boldsymbol\beta$ \Comment{Change the weights for layers of the discriminator network} 	\State Compute $\nabla_\btheta L_D^{(j)}(\tilde{\btheta})$ 	
	\EndFor 	
	\State  $\btheta=\btheta - \eta_t\times\bigg(\sum_j\nabla_\btheta L_D^{(j)}(\tilde{\btheta})\bigg)$ \Comment{Alternatively use Equation \ref{eq:langevin_update}} 	
	\State Normalize $\btheta$ so that $\|\btheta\|\leq 1$ 	\For{$j=1\to m'$} 
	\State $S'=$\Call{SampleFake}{$\bomega$} 		
	\State {Compute $\nabla_\bomega L_G^{(j)}(\tilde{\bomega})$} 	
	\EndFor 	
\State $\bomega=\bomega - \eta_t\times\bigg(\sum_j\nabla_\bomega L_G^{(j)}(\tilde{\bomega})\bigg)$ \Comment{Alternatively use Equation \ref{eq:langevin_update}} 
\EndWhile 
\State \Return $\btheta, \bomega$ 
\Statex 
\Procedure {SampleFake}{$\bomega$} 	
\State{Sample $\boldsymbol{\alpha}\sim\text{Bernoulli}{(\pi_G)}$} 	
\State{Sample $\boldsymbol{\beta}\sim\mathcal{N}(0, \sigma_G^2\mathbf{I})$}   \Comment{According to dimenstions of $\btheta$} 	
\State $\tilde{\bomega}=\bomega\odot\boldsymbol\alpha+\boldsymbol\beta$ \Comment{Change the weights for layers of the generator network} 	
\State{$S'$ = Sample a batch $\{(\bx'_i,y'_i)\}_{i=1}^{n'}$ from the generator network using $\tilde{\bomega}$} 	
\State\Return {$S'$} 
\EndProcedure \end{algorithmic}

\caption{Our Bayesian GAN algorithm.}
\end{algorithm}

\section{Sampling Functions}

At each step of our algorithm we need to compute expectations with
respect to the generator and discriminators. We do this by taking
samples of each function according to their distributions. This is
done using simple tricks like dropout that allow us to sample from
a neural network. It is shown that dropout \cite{SrivastavaHintonKrizhevskyEtAl2014}
has a Bayesian interpretation where the posterior is approximated
using variational inference \cite{GalGhahramani2015}. The connection
between Gaussian Processes \cite{MacKay2003,RasmussenWilliams2006}
and dropout for classification is made by placing a variational distribution
over variables in the model and minimizing the KL-divergence between
this variational distribution and the true distribution of the variables.
Dropout acts as a regularizer too and improves the generalization
performance of neural nets as reported in \cite{SrivastavaHintonKrizhevskyEtAl2014}.
As such, we use dropout as a means of sampling various functions for
the generator and discriminator. 

For the discriminator, we use variants of dropout for estimating the
uncertainty of the discriminator in its predictions using the variance
of the predictive distribution: {\small{}
\begin{eqnarray*}
p(y^{*}|\bx^{*},\btheta) & = & \int p(f_{D}(\bx^{*}),y^{*})p(f_{D}|\btheta)df_{D}\\
\mathbb{E}[y^{*\top}y^{*}|\bx^{*}] & \approx & \tau^{-1}\mathbf{I}+\frac{1}{m}\sum p(f_{D}(\bx^{*}){}^{\top}f_{D}(\bx^{*}))\\
\mathbb{V}[y^{*}|\bx^{*}] & \approx & \mathbb{E}[y^{*\top}y^{*}|\bx^{*}]-\mathbb{E}[y^{*}|\bx^{*}]^{2}\qquad\qquad f_{D}=a(.,\tilde{{\btheta}}),\tilde{\btheta}=\btheta\odot\boldsymbol{\alpha}+\boldsymbol{\beta}
\end{eqnarray*}
}where $a$ denotes the activation function (we slightly misused the
notation for indication of the predictive mean and variance), $\tau>0$
is the variance of the prior of the weights and $\mathbf{I}$ is the
identity matrix. Here, $\bx^{*},y^{*}$ denote test instance and its
corresponding predicted label either from the real or fake dataset.
We use the same trick of using variants of dropout to obtain samples
of the generator function too.

While GPs define distributions over functions in a non-parametric
Bayesian manner by analytically integrating out the parameters, we
sample the parameters and use the Monte Carlo method to estimate their
expectation. 

\section{Choice of discrepancy measure}

When the generator network generates high quality fake samples, the
discrepancy between the fake samples and real samples is expected
to be small. A suitable discrepancy measure should capture statistical
properties of the real and fake data. We choose the \emph{maximum
mean discrepancy} (MMD) measure which asserts when the dimensions
of the data is large and the moment matching in the input space is
not possible, difference between the empirical means of two distributions
using a nonlinear feature map is a measure of closeness for two-sample
problems. The feature map that used in this measure has to be bounded
and compact. This property is ensured by constraining the weights,
i.e. $\|\btheta\|^{2}\leq1$. Our $\Delta$ function is defined as,

{\scriptsize{}
\begin{eqnarray*}
\Delta_{f_{D}^{(i)}}(S,S') & = & \bigg\|\frac{1}{n}\sum_{l}f_{D}^{(i)}(\bx_{l})-\frac{1}{n'}\sum_{l'}f_{D}^{(i)}(\bx_{l'}')\bigg\|^{2},\qquad\qquad\|\boldsymbol{\theta}\|\leq1
\end{eqnarray*}
}{\scriptsize \par}

It is interesting to note that in the neural network implementation
of this measure, we only need to ensure the parameters are normalized.
The value of weight normalization in neural nets have already been
shown in \cite{SalimansKingma2016}. Here we show it can further be
used in a different manner in our GAN model for density comparison.


\section{Experiments}

\begin{wrapfigure}{r}{0.4\textwidth}   
\begin{center}     
\vspace{-15mm}
\subfigure[]{\includegraphics[width=0.15\textwidth]{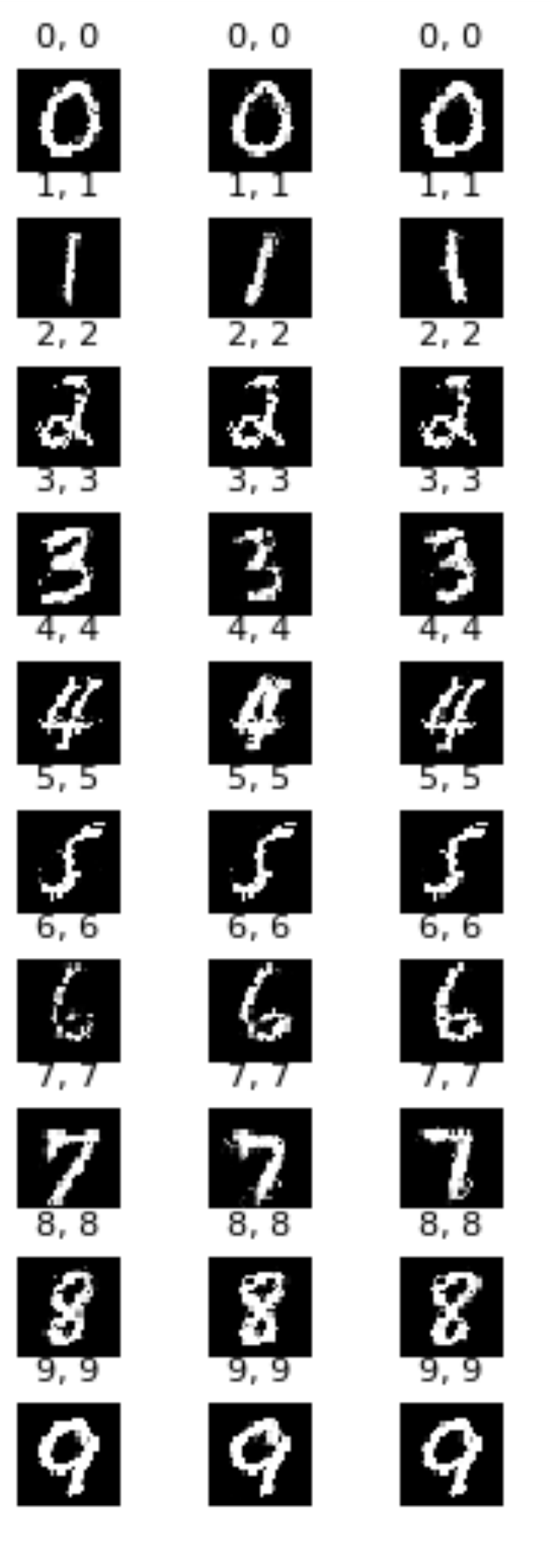}\label{fig:mnist_1}}
\hspace{6mm}
\subfigure[]{\includegraphics[width=0.15\textwidth]{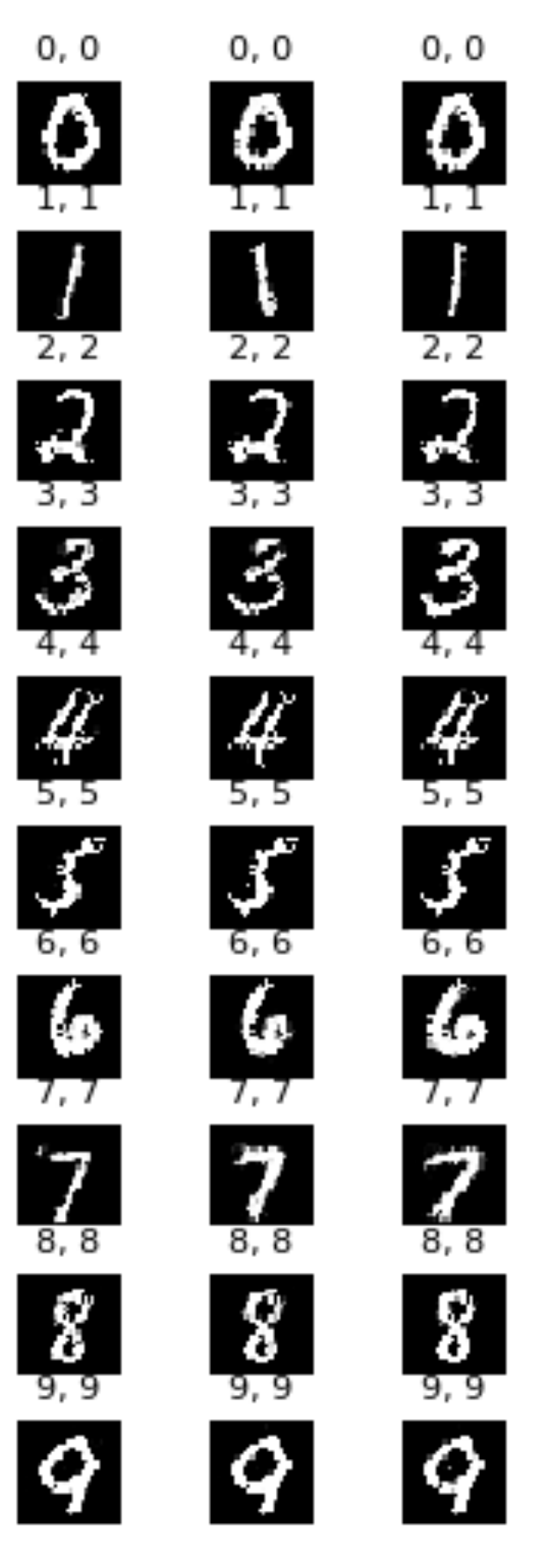}\label{fig:mnist_2}}
\end{center}   
\vspace{0mm}
\caption{Samples of the conditional network in MNIST dataset using MAP-MC in Figure \ref{fig:mnist_1} Langevin dynamics in Figure \ref{fig:mnist_2}.} 
\label{fig:mnist_samples}
\end{wrapfigure}

Evaluation of generative models in general and GANs in particular
is typically difficult. Since our approach has a classification loss
as well, we target semi-supervised learning problems. In particular
we can use small number of training instances with $\log$ loss and
a softmax layer for the output of the discriminator for training our
model. We also observed it is important to add an output for fake
images in the loss (i.e. have $K+1$ output for the discriminator
network). We use a one-hot vector of labels for the input to the generator.
This deterministic input may cause collapse for the whole generator
network especially if the randomizations are not enough. We add layers
of dropout and Gaussian noise to every layer from the input. For all
the experiments we set the batch sizes for stochastic gradient descent
to $100$. We randomly select a subset of labeled examples and use
the rest as unlabelled data for training. We perform these experiments
$5$ times and report the mean and standard error. We use a constant
learning rate for MAP-MC approach and reduce the learning rate with
inversely proportionate rate with the training epoch for the Langevin
dynamics. For the Monte Carlo samplings we use only $2$ samples for
efficiency.

We evaluate our approach on two datasets: MNIST and CIFAR-10. In our
experiments, MAP-MC performs better than Langevin dynamics in terms
of accuracy that we conjecture is due to the nature of the inference
method. Intuitively in Langevin dynamics, the gradients are noisy
and thus the movement of the parameters in such a complex space with
minimax objective is difficult.

\begin{table}
\centering{}%
\begin{tabular}{|c|c|c|c|}
\hline 
{\footnotesize{}{}Method/Labels}  & {\footnotesize{}{}100}  & {\footnotesize{}{}1000}  & {\footnotesize{}{}All}\tabularnewline
\hline 
\hline 
{\footnotesize{}{}Pseudo-label \cite{Lee2013}}  & {\footnotesize{}{}$10.49$}  & {\footnotesize{}{}$3.64$}  & {\footnotesize{}{}$0.81$}\tabularnewline
\hline 
{\footnotesize{}{}DGN \cite{KingmaMohamedRezendeEtAl2014}}  & {\footnotesize{}{}$3.33\pm0.14$}  & {\footnotesize{}{}$2.40\pm0.02$}  & {\footnotesize{}{}$0.96$}\tabularnewline
\hline 
{\footnotesize{}{}Adversarial \cite{GoodfellowPouget-AbadieMirzaEtAl2014}}  &  &  & {\footnotesize{}{}$0.78$}\tabularnewline
\hline 
{\footnotesize{}{}Virtual Adversarial \cite{MiyatoMaedaKoyamaEtAl2016}}  & {\footnotesize{}{}$2.66$}  & {\footnotesize{}{}$1.50$}  & {\footnotesize{}{}$0.64\pm0.03$}\tabularnewline
\hline 
{\footnotesize{}{}PEA \cite{BachmanAlsharifPrecup2014}}  & {\footnotesize{}{}$5.21$}  & {\footnotesize{}{}$2.64$}  & {\footnotesize{}{}$2.30$}\tabularnewline
\hline 
{\footnotesize{}{}$\Gamma\text{-Model}$ \cite{RasmusValpolaHonkalaEtAl2015}}  & {\footnotesize{}{}$4.34\pm2.31$}  & {\footnotesize{}{}$1.71\pm0.07$}  & {\footnotesize{}{}$0.79\pm0.05$}\tabularnewline
\hline 
\hline 
{\footnotesize{}BC-GAN (MAP-MC)} & {\footnotesize{}{}$1.01\pm0.05$}  & \textbf{\footnotesize{}{}$0.86\pm0.04$}{\footnotesize{} } & \textbf{\footnotesize{}{}$0.7\pm0.03$}\tabularnewline
\hline 
{\footnotesize{}BC-GAN (Langevin)} & {\footnotesize{}$2.5\pm0.5$}  & \textbf{\footnotesize{}$1.6\pm0.9$}{\footnotesize{} } & \textbf{\footnotesize{}$0.8\pm0.09$}\tabularnewline
\hline 
\end{tabular}\caption{Semi-supervised learning using our approach compared to others on
MNIST dataset. As shown approach is comparable or better than its
counterparts.}
\vspace{-5mm}\label{tbl:mnist_semi} 
\end{table}

\textbf{MNIST Dataset:} MNIST dataset\footnote{http://yann.lecun.com/exdb/mnist/}
contains $60,000$ training and $10,000$ test images of size $28\times28$
of handwritten digits as greyscale images. Since the image pixels
are binary, we use a generator network with sigmoid activation for
the output of each pixel. For generator, we sample the one-hot label
vector $y'$ for the input uniformly for all classes (for 10 classes
we have equal number of samples as the input for generator net). We
use a three layer generator network with $500$ softplus activation
units. In between these fully connected layers, we use Gaussian and
Bernoulli dropout with variance $0.9$ and ratio $0.1$ respectively.
We also used batch normalization after each layer as was shown to
be effective \cite{Springenberg2015}. \begin{wrapfigure}{l}{0.4\textwidth}   
\begin{center}     
\vspace{0mm}
\includegraphics[width=0.38\textwidth]{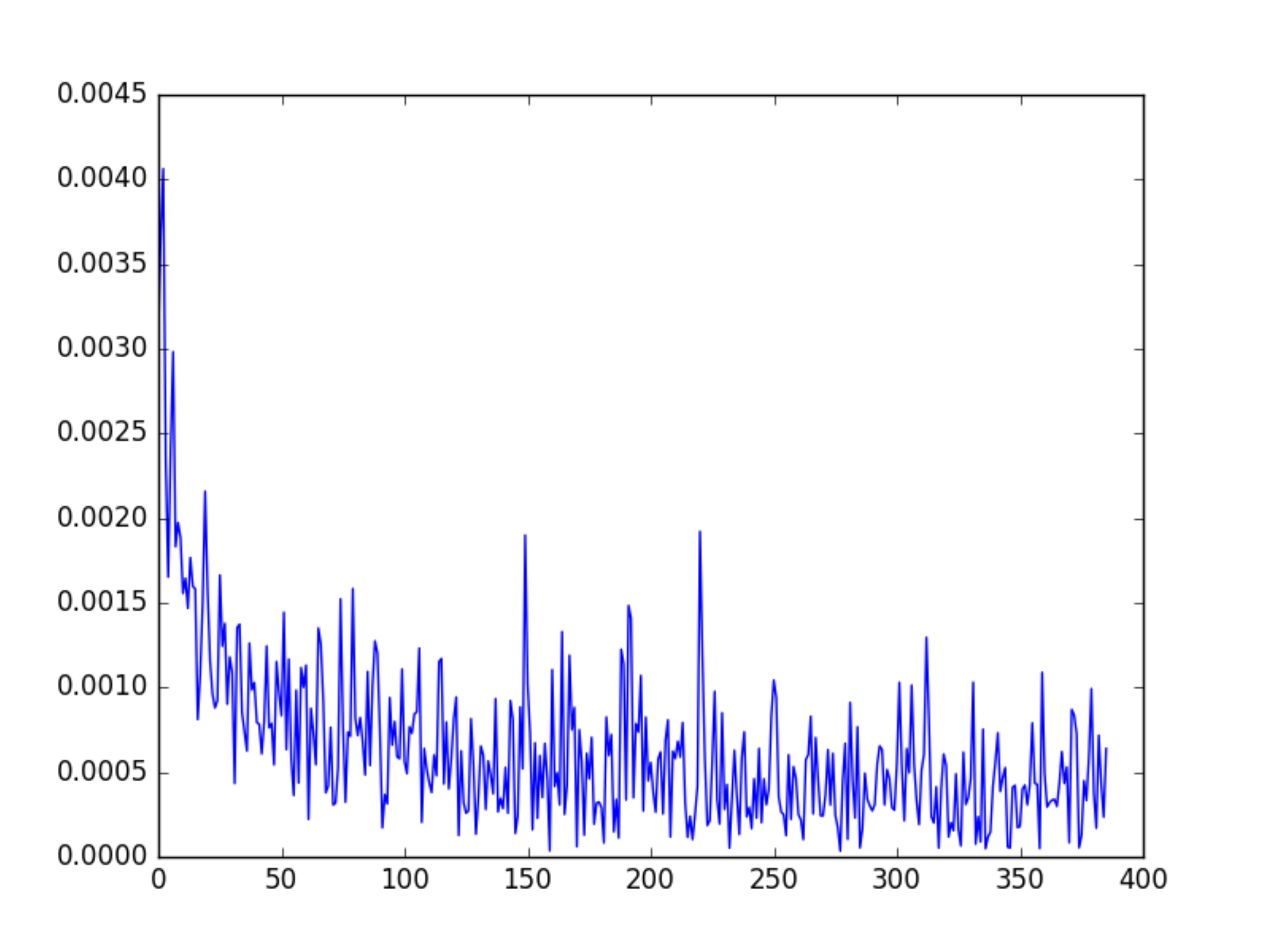}
\end{center}   
\vspace{-5mm}
\caption{Predictive variance of the discriminator on 10 randomly selected samples during training in MNIST dataset using MAP-MC. In the x-axis we show the epoch and in y-axis the variance. As seen, the variance is reduced with training.} 
\label{fig:mnist_var}
\vspace{-3mm}
\end{wrapfigure}

The discriminator is a fully connected network with Gaussian and Bernoulli
dropout layers in between with variance $0.9$ and ratio $0.05$ respectively.
We use weight normalization at the last layer (using it in all the
layers seemed to improve convergence speed). As shown in Figure \ref{fig:mnist_samples},
we can generate samples of real looking images from the MNIST dataset
for each label. The small numbers above generated images are the generated
label and the discriminator's prediction respectively. As observed,
at the final stages of training discriminator is so powerful that
can basically predict almost all generated labels correctly. Generator
is also trained to match the generated class with the corresponding
image. It should be noted that the samples here exploit the uncertainty
of the generator network function which is desirable. Test errors
for semi-supervised learning using our approach compared to the state
of the art is presented in Table \ref{tbl:mnist_semi}. Furthermore,
predictive variance of the discriminator as a measure of its uncertainty
when using only 10 labelled instances for each class is shown in Figure
\ref{fig:mnist_var}. As expected with training, the variance is reduced
and the discriminator becomes more confident about its predictions.

\textbf{CIFAR-10 Dataset}: The CIFAR-10 dataset \cite{KrizhevskyHinton2009}
is composed of 10 classes of natural $32\times32$ RGB images with
$50,000$ images for training and $10,000$ images for testing. Complexity
of the images due to higher dimensions, color and variability make
this task harder. We use a fully connected layer after the input and
three layers of deconvolution that are batch normalized to generate
samples. Again, we use Bernoulli and Gaussian dropout between layers
with $0.9$ and ratio $0.2$ to induce uncertainty. 

\begin{figure}
\centering
\subfigure{\hspace{-10mm}\includegraphics[width=220pt,trim={35mm 15mm 15mm 5mm},clip]{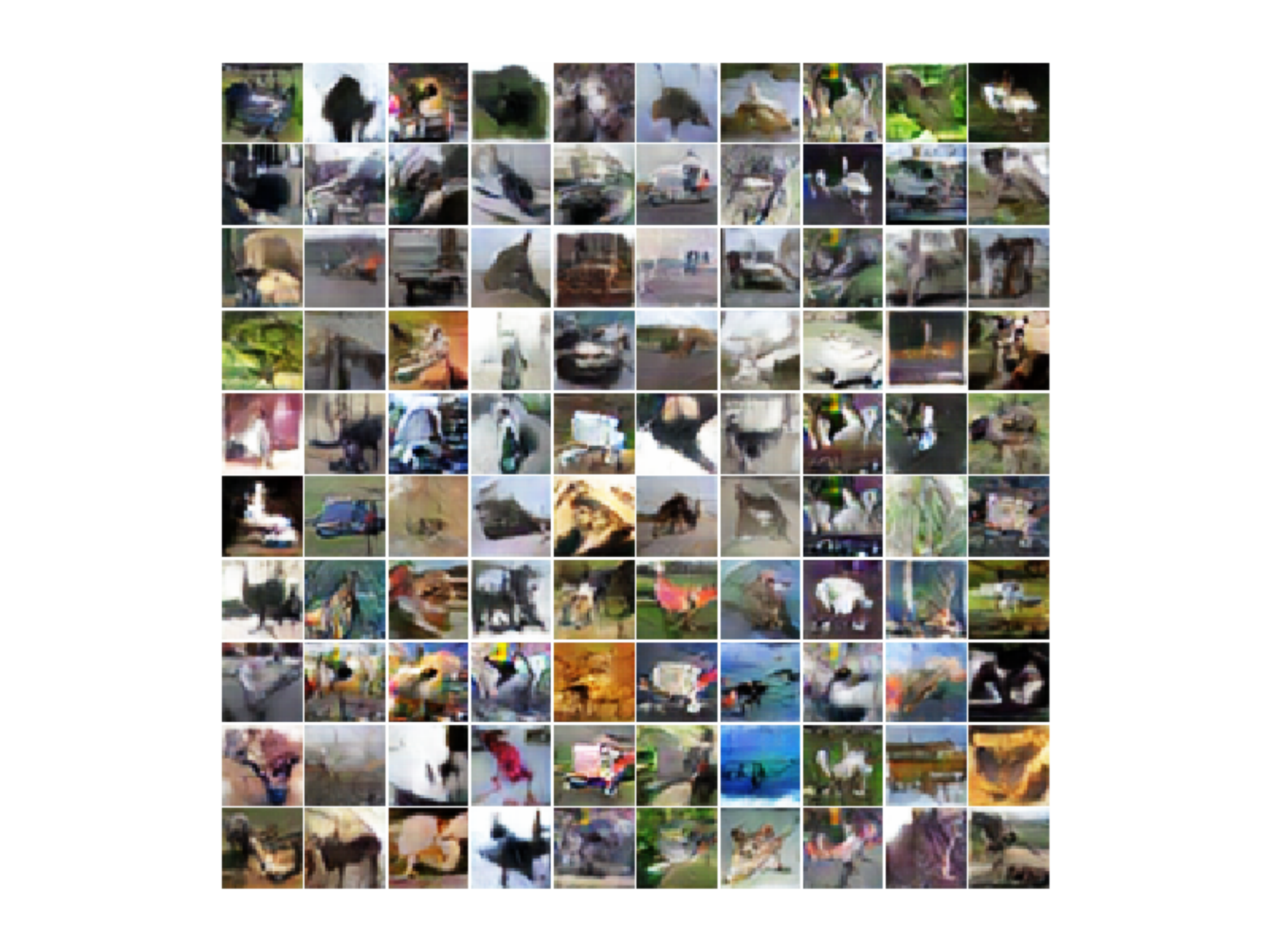}}\hspace{-5mm}\subfigure{\includegraphics[width=220pt,trim={35mm 15mm 15mm 5mm},clip]{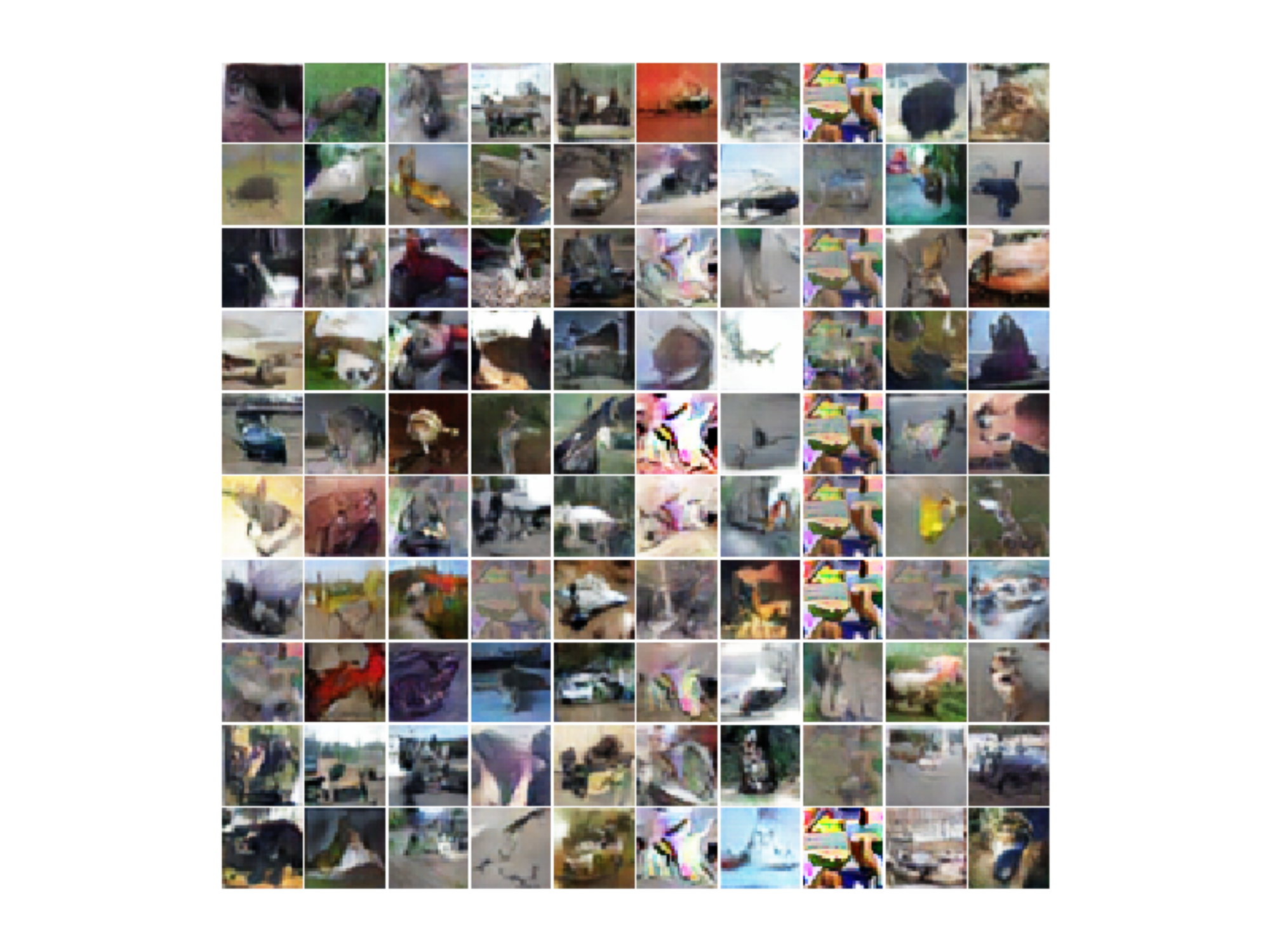}\hspace{-20mm}}\vspace{-3mm}

\caption{Samples of generated images from the CIFAR-10 dataset.}
\vspace{-4mm}\label{fig:cifar}
\end{figure}

For the discriminator network we use a 9 layers convolutional net
with two fully connected layers in the output. Furthermore, we use
weight normalization at each layer. We report the performance of our
approach on this dataset for semi-supervised learning in Table \ref{tbl:cifar10}.
There are samples from the generator shown in Figure \ref{fig:cifar}.
As observed in case of MAP-MC, we have diverse images with similarity
across columns where images have same label. Langevin approach on
the other hand, suffers from the mode collapse in one of the classes
where some of the images seem similar (third column from left). It
suggests we should have stopped earlier for the Langevin or used higher
variance or ratio in dropout. 

\begin{table}
\centering{}{\footnotesize{}}%
\begin{tabular}{|c|c|c|c|}
\hline 
{\footnotesize{}{}Method/Labels } & {\footnotesize{}{}1000 } & {\footnotesize{}{}4000 } & {\footnotesize{}{}All}\tabularnewline
\hline 
\hline 
{\footnotesize{}{}Conv-Large \cite{RasmusValpolaHonkalaEtAl2015,SpringenbergDosovitskiyBroxEtAl2014} } &  & {\footnotesize{}{}$23.3\pm30.61$ } & {\footnotesize{}{}$9.27$}\tabularnewline
\hline 
{\footnotesize{}{}$\Gamma\text{-Model}$ \cite{RasmusValpolaHonkalaEtAl2015} } &  & {\footnotesize{}{}$20.09\pm0.46$ } & {\footnotesize{}{}$9.27$}\tabularnewline
\hline 
{\footnotesize{}CatGAN \cite{Springenberg2015}} &  & {\footnotesize{}$19.58\pm0.46$} & \tabularnewline
\hline 
\hline 
{\footnotesize{}Bayesian-GAN using MAP-MC } & {\footnotesize{}{}$20.9\pm1.05$ } & {\footnotesize{}{}$18.89\pm0.65$ } & {\footnotesize{}{}$7.9\pm0.3$}\tabularnewline
\hline 
{\footnotesize{}Bayesian-GAN using Langevin dynamics} & {\footnotesize{}$25.8\pm1.45$ } & {\footnotesize{}$20.1\pm1.45$ } & {\footnotesize{}$9.3\pm0.8$ }\tabularnewline
\hline 
\end{tabular}\caption{Test error on CIFAR10 with various number of labeled training examples. }
\label{tbl:cifar10} \vspace{-5mm}
\end{table}

\section{Related work}

Since inception of GANs \cite{GoodfellowPouget-AbadieMirzaEtAl2014}
for unsupervised learning, various attempts have been made in either
better explaining these models, extending them beyond unsupervised
learning or generating more realistic samples. Wasserstein GAN \cite{ArjovskyChintalaBottou2017}
and Loss-Sensitive GAN \cite{Qi2017} are recently developed theoretically
grounded approaches that provide an information-theoretic view of
these models. The objective in these methods are constructed to be
more generic than binary classification done in the discriminator.
In fact, discriminator acts as a measure of closeness for the generated
samples and the real ones similar approach that is taken in this paper.

Further, GANs have been successfully used for latent variable modeling
and semi-supervised learning with the intuition that the generator
assists the discriminator when the number of labelled instances are
small. For instance, InfoGAN \cite{NIPS2016_6399} proposed to learn
a latent variable that represents cluster of data while learning to
generate images by utilizing variational inference. While it was not
directly used for semi-supervised learning, its extension categorical
GAN (CatGAN) {\footnotesize{}\cite{Springenberg2015}} that utilized
mutual information as part of its loss has developed with very good
performance. Furthermore, \cite{SalimansGoodfellowZarembaEtAl2016}
have developed heuristics for better training and achieving state
of the art results in semi-supervised learning using GANs. 

On the other hand, unlike our approach conditional GAN \cite{Mirza2014arXiv1411.1784M}
generate labelled images by adding the label vector to the input noise.
Furthermore, MMD measure \cite{gretton2012kernel} have been used
in GANs with success \cite{2016arXiv161104488S}. However previously,
kernel function had to be explicitly defined and in our approach we
learn it as part of the discriminator. 

Use of Bayesian methods in GANs have generally been limited to combining
variational autoencoders \cite{KingmaWelling2014} with GANs \cite{2015arXiv151209300B,dumoulin2017adversarially-iclr}.
We on the other hand take a Bayesian view on both discriminator and
generator using the dropout approximation for variational inference.
This allows us to develop a simpler and more intuitive model.

\section{Conclusion}

Unlike traditional GANs, we proposed a conditional Bayesian model,
called BC-GAN, that exploits the uncertainty in the generator as a
source of randomness for generating real samples. Similarly, we evaluate
the uncertainty of the discriminator that can be used as a measure
of its performance. This allows us to better analyze the behavior
of a good generator/discriminator from a functional perspective for
future and shed light on the GANs. We also evaluated our approach
in a semi-supervised learning problem and showed its effectiveness
in achieving state of the art results.

In future we plan to explore other inference methods such as Hamiltonian
Monte Carlo that may yield better performance. In addition, we hope
to perform more analysis on our method to better explain its internal
behavior in a minimax model.

\footnotesize{\bibliographystyle{natbib}
\bibliography{library.bib}
}
\end{document}